\documentclass[lettersize,journal]{IEEEtran}
\usepackage{amsmath,amsfonts}
\usepackage{algorithmic}
\usepackage{algorithm}
\usepackage{array}
\usepackage[caption=false,font=normalsize,labelfont=sf,textfont=sf]{subfig}
\usepackage{textcomp}
\usepackage{stfloats}
\usepackage{url}
\usepackage{verbatim}
\usepackage{graphicx}
\usepackage{cite}
\usepackage{multirow}
\usepackage{booktabs}

\begin{document}

\title{NM-FlowGAN: Modeling sRGB Noise without Paired Images using a Hybrid Approach of Normalizing Flows and GAN} 

\author{
  Young~Joo~Han,
  Ha-Jin~Yu,~\IEEEmembership{Member,~IEEE}%
  \thanks{This work was supported by the 2024 sabbatical year research grant of the University of Seoul. (Corresponding author: Ha-Jin~Yu.)}
  \thanks{
    Y.~J.~Han and H.~J.~Yu are with the
    School of Computer Science, University of Seoul, Seoul 02504, South Korea
    (e-mail: orora71@gmail.com;hjyu@uos.ac.kr).}%
  \thanks{
    Y.~J.~Han is with the
    R\&D Center, Vieworks, Anyang-si 14055, South Korea}%
  }



\maketitle

\begin{abstract}
Modeling and synthesizing real sRGB noise is crucial for various low-level vision tasks, such as building datasets for training image denoising systems. The distribution of real sRGB noise is highly complex and affected by a multitude of factors, making its accurate modeling extremely challenging. Therefore, recent studies have proposed methods that employ data-driven generative models, such as Generative Adversarial Networks (GAN) and Normalizing Flows. These studies achieve more accurate modeling of sRGB noise compared to traditional noise modeling methods. However, there are performance limitations due to the inherent characteristics of each generative model. To address this issue, we propose NM-FlowGAN, a hybrid approach that exploits the strengths of both GAN and Normalizing Flows. We combine pixel-wise noise modeling networks based on Normalizing Flows and spatial correlation modeling networks based on GAN. Specifically, the pixel-wise noise modeling network leverages the high training stability of Normalizing Flows to capture noise characteristics that are affected by a multitude of factors, and the spatial correlation networks efficiently model pixel-to-pixel relationships. In particular, unlike recent methods that rely on paired noisy images, our method synthesizes noise using clean images and factors that affect noise characteristics, such as easily obtainable parameters like camera type and ISO settings, making it applicable to various fields where obtaining noisy-clean image pairs is not feasible. In our experiments, our NM-FlowGAN outperforms other baselines in the sRGB noise synthesis task. Moreover, the denoising neural network trained with synthesized image pairs from our model shows superior performance compared to other baselines.
\end{abstract}


\section{Introduction}
\label{sec:intro}

\IEEEPARstart{R}{eal-world} image denoising is a fundamental task in low-level vision. However, training effective neural networks for real-world image denoising is highly challenging, due to the difficulty in obtaining noisy-clean image pairs for training image denoising neural networks. A noisy-clean image pair consists of a noisy image and a corresponding clean image that shares the same noise-free content. Because noise is affected by factors such as lighting conditions, sensor limitations, or environmental influences, achieving the perfect alignment between noisy and clean images is a significant challenge.
Therefore, existing real-world image denoising datasets have been limited to those captured in controlled environments, where stationary imaging devices are used to capture static scenes~\cite{sidd, nind, polyu}. Consequently, image denoising neural networks trained solely on data obtained under such conditions may exhibit degraded performance when processing images from uncontrolled environments, such as dynamic scenes.

\begin{figure}[!t]
    \centering
    \renewcommand{\wp}{0.95\linewidth}
    \captionsetup[subfloat]{font=scriptsize} 
    \centering
    \subfloat[Training noise modeling networks]
    {\includegraphics[width=\wp]{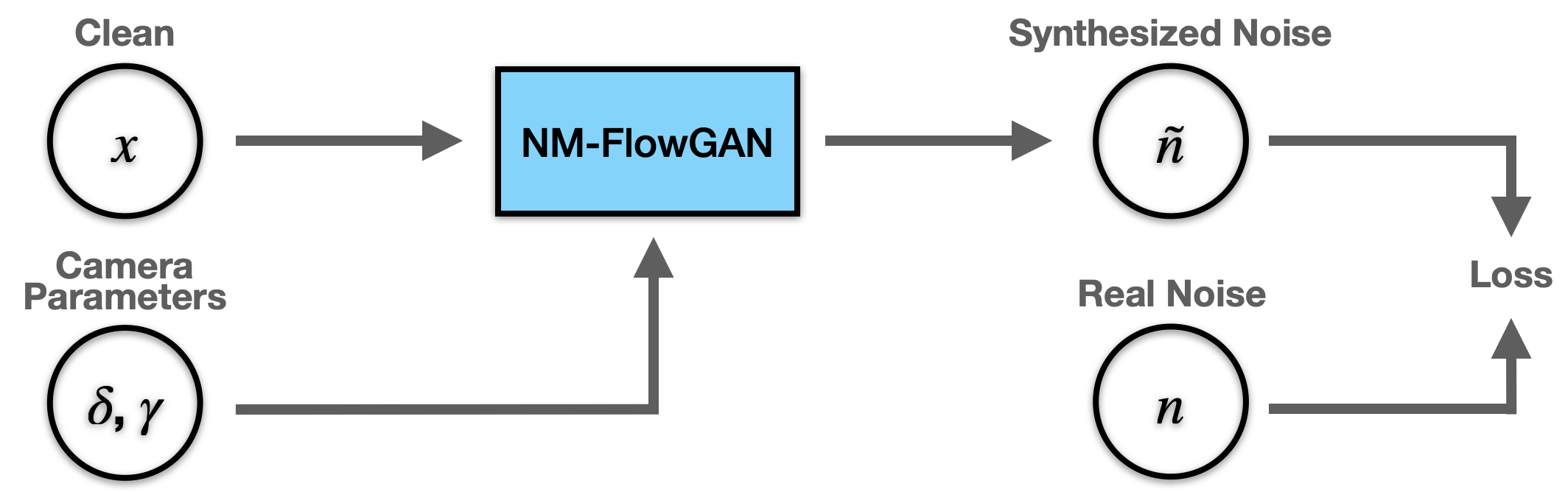}\label{fig:fig_overview_noisemodeling}} 
    \hfill
    \subfloat[Training image denoising networks]
    {\includegraphics[width=\wp]{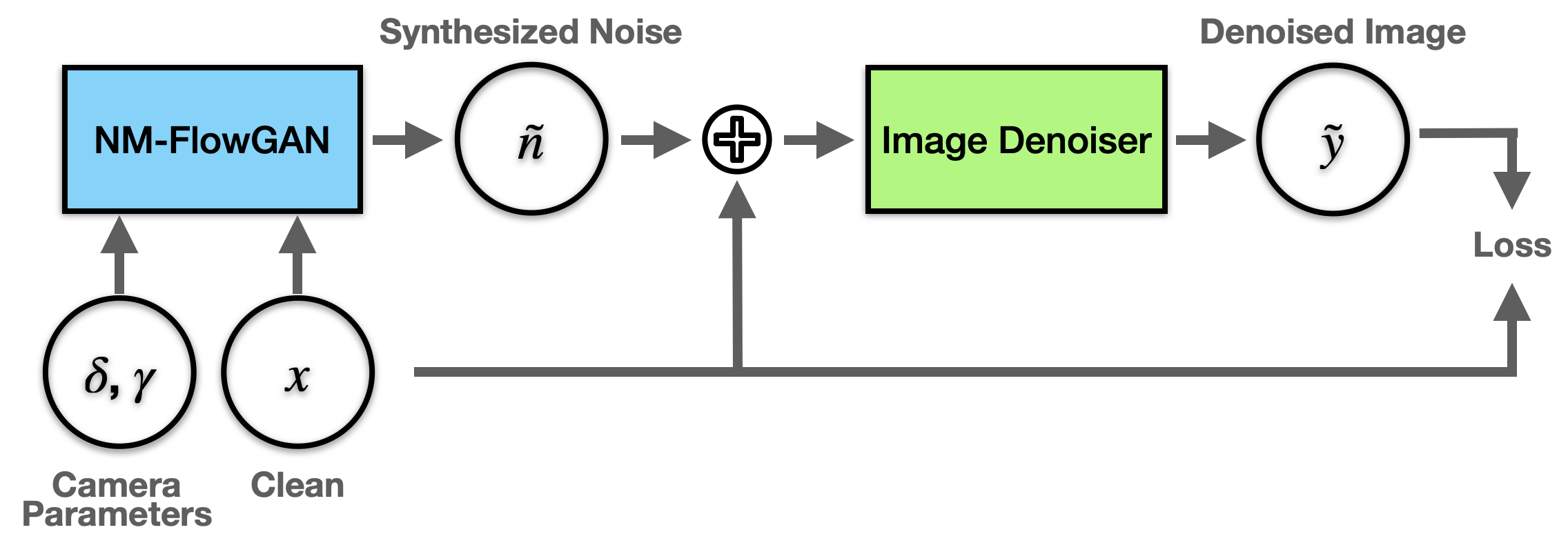}\label{fig:fig_overview_denoiser}} %
    \hfill

    \caption{
        \textbf{Two-step pipeline for training real-world image denoising networks with noise modeling networks.} (a) First, our method trains noise modeling networks called NM-FlowGAN to synthesize noise $\tilde{n}$ using clean images $x$ and camera parameters $\delta$ and $\gamma$, which represent the camera type and ISO setting. (b) Once trained, NM-FlowGAN synthesizes noise that is added to clean images, generating synthesized noisy-clean image pairs that are used to train image denoising networks.}
   \label{fig:fig_overview}
\end{figure}
\IEEEpubidadjcol

One effective approach to overcome this challenge is noise modeling, which involves estimating noise characteristics of imaging devices such as smartphone cameras. Specifically, noise modeling neural networks can be trained using data captured in controlled environments to model the noise from imaging devices. Once trained, these noise modeling neural networks can synthesize noise into clean images captured in various environments, generating a large amount of noisy-clean image pairs for training image denoising networks. This method can generate noisy-clean image pairs even in uncontrolled environments where obtaining noisy-clean image pairs is not feasible due to physical constraints. Consequently, image denoising neural networks trained on such synthesized noisy-clean image pairs can demonstrate robust performance on scenes captured in various environments. Figure~\ref{fig:fig_overview} illustrates the process of training a noise modeling network and using it to train image denoising neural networks.
Furthermore, noise modeling can also be applied to medical imaging~\cite{nm_x_ray}, such as X-ray imaging, where the acquisition of noisy-clean image pairs is particularly challenging. In X-ray imaging, capturing multiple images of the same scene can result in unnecessary radiation exposure, raising serious health concerns for patients. Additionally, involuntary movements of internal organs make it difficult to acquire perfectly aligned image pairs, which are essential for supervised learning in denoising tasks. By utilizing noise modeling techniques that rely solely on clean images and known imaging parameters, we can generate synthetic noisy images without subjecting patients to additional radiation or relying on impractical imaging conditions. This approach not only preserves patient safety but also facilitates the development of more effective denoising algorithms for medical applications.
Due to these advantages, numerous studies have focused on accurate noise modeling. The simplest and most commonly used noise models include additive white Gaussian noise (AWGN) and Poisson noise. However, these simple models are insufficient to capture the complex distribution of real-world noise. To address this issue, more advanced models like the Poisson-Gaussian noise model~\cite{foi2008practical} and the heteroscedastic Gaussian noise model~\cite{foi2009clipped, sidd} have been proposed. While these methods have shown improved performance over simpler models, they still do not fully encapsulate the intricate characteristics of real-world noise.
Recently, numerous data-driven methods~\cite{noise_flow, ca_noise_gan, cbdnet, noise2noiseflow, gcbd, song2023generative} based on deep learning have been proposed. These methods are more sophisticated than previous ones, capable of representing more complex distributions. They use neural networks trained on large-scale datasets of noisy images to learn the noise distribution, focusing on modeling both signal-dependent and signal-independent noise from raw sensor images. These approaches can model noise much more accurately, and the denoising neural networks trained on the modeled noise also show promising performance.
Building on the success of noise modeling from raw sensor images, studies have been proposed to perform noise modeling in the standard RGB (sRGB) color space~\cite{srgb_flow, neca, c2n, noisetransfer, song2023generative}. Generally, noise modeling is categorized based on the image domain in which it is conducted, either in the raw-RGB or the sRGB domains. Images in the raw-RGB domain are unprocessed raw sensor data without the application of image signal processing (ISP), maintaining the original data captured by the sensor. In contrast, images in the sRGB domain have undergone the ISP, resulting in standardized and enhanced visuals suitable for various displays and environments.
Noise modeling is notably more difficult in the sRGB domain than in the raw-RGB domain due to the ISP's impact on both clean signals and noise, resulting in a complex noise distribution~\cite{holistic}. Nevertheless, noise modeling in the sRGB domain is essential because the conversion from raw-RGB to sRGB is performed by in-camera processing~\cite{srgb_flow}. Often, image denoising is not applied or is insufficiently applied prior to the ISP. Consequently, effective noise modeling in the sRGB domain becomes necessary.
Most deep learning-based noise modeling methods in the sRGB domain employ generative models because they can learn complex, high-dimensional data distributions, which is essential for accurately capturing the statistical properties of noise. Efforts have been made to model sRGB noise using generative models such as Generative Adversarial Networks (GAN)~~\cite{gan} and Normalizing Flows~~\cite{nf, nf2}. These methods have shown promising performance in sRGB noise modeling; however, each of these generative models has its own strengths and weaknesses, limiting their effectiveness.
GAN-based methods~\cite{neca, c2n, noisetransfer} are well-suited to handling high-dimensional and complex data distributions, enabling them to generate realistic noise images in the sRGB domain. However, they have limitations in training stability when dealing with small-sized datasets. Noise modeling often involves insufficient datasets because generating noisy-clean pairs is highly time-consuming. Additionally, the noise distribution is affected by various camera conditions (e.g., types, settings)~\cite{sidd, nind}. Therefore, it is crucial to employ a generative model that can maintain high training stability even with limited data for each specific condition.
In contrast, Normalizing Flows-based methods~\cite{srgb_flow, naflow} exhibit higher training stability with small datasets by directly learning the probability density function of the data. However, due to the constraints of Normalizing Flows, where each layer must be invertible, there are limitations on the expressiveness of the transformation functions. Consequently, generating realistic noise images remains challenging compared to GAN-based methods.
To address the limitations of both GAN-based and Normalizing Flows-based methods, we propose a novel data-driven sRGB noise modeling method named NM-FlowGAN, which is a hybrid approach that combines the strengths of Normalizing Flows and GAN. Our method is designed to effectively model the highly complex noise distribution found in the sRGB image domain. By integrating the high training stability of Normalizing Flows with the powerful expressiveness of GAN, we can overcome the individual limitations of each method.
Moreover, unlike recently proposed methods such as NeCA-W~\cite{neca} and NAFlow~\cite{naflow}, which have demonstrated promising noise modeling performance but rely on real noisy images paired with clean images\footnote{This implies that additional synthesized noisy-clean image pairs are generated from the original noisy-clean pair.}, our method synthesizes noisy images without such pairs. Using real paired noisy images enables the generation of realistic noise and can enhance the performance of image denoising neural networks through data augmentation. However,  the requirement for paired real noisy images to synthesize noise makes it impossible to apply these methods in environments where obtaining noisy-clean image pairs is not feasible. By relying only on clean images and factors that affect noise characteristics, such as camera type or ISO settings, our method broadens the applicability of noise modeling to various environments such as dynamic scenes.
In our proposed method, we first take advantage of the high training stability offered by Normalizing Flows to model pixel-wise noise. Our analysis of noise in the sRGB domain reveals that pixel-wise noise depends on the clean image intensity, surrounding image structures, and camera conditions. To address this, we design novel invertible layers that can effectively model these dependencies. Additionally, we employ GAN to model high-dimensional noise features, such as spatial correlations induced by in-camera imaging processes like demosaicing. This combined approach enables the generation of realistic sRGB noise images while ensuring high training stability. To validate our method, we compared the performance of noise modeling in the sRGB image domain in real-world scenarios. In the experiments, our method shows superior performance compared to the other baselines. 

\section{Related Works}
\label{sec:related}

\subsection{Generative Adversarial Network}

\noindent GAN has been actively adopted for modeling complex noise that occurs in real-world scenarios. There are a few efforts to apply this for modeling noise in the sRGB image domain. Firstly, C2N~\cite{c2n} attempts to model noise in the sRGB image domain using unpaired noisy-clean image pairs. However, because they do not consider camera conditions that affect the noise distribution, the noise images generated from the model suffer from color shift problems. 
To address the issue, a noise modeling method NeCA~\cite{neca} is proposed, which estimates the noise level using an additional noisy image and models spatially correlated noise~\cite{Zhou} using GAN. While this method can generate realistic noise images, its performance is significantly affected by the noisy image used in gain estimation. In addition, the method has the problem that it requires extensive training time due to the necessity of training neural networks for each camera type.
Consequently, GAN-based methods are highly effective in modeling high-dimensional noise features and generating realistic noise images in the sRGB image domain. However, they have limitations in considering camera conditions that affect noise distribution due to the problem of training stability on small-sized datasets.
\begin{figure*}[!t]
    \centering
    \renewcommand{\wp}{0.333\linewidth}
    \captionsetup[subfloat]{font=scriptsize} 
    \centering
    \subfloat[Scene: 8, Cam: G4, ISO: 100]
    {\includegraphics[width=\wp]{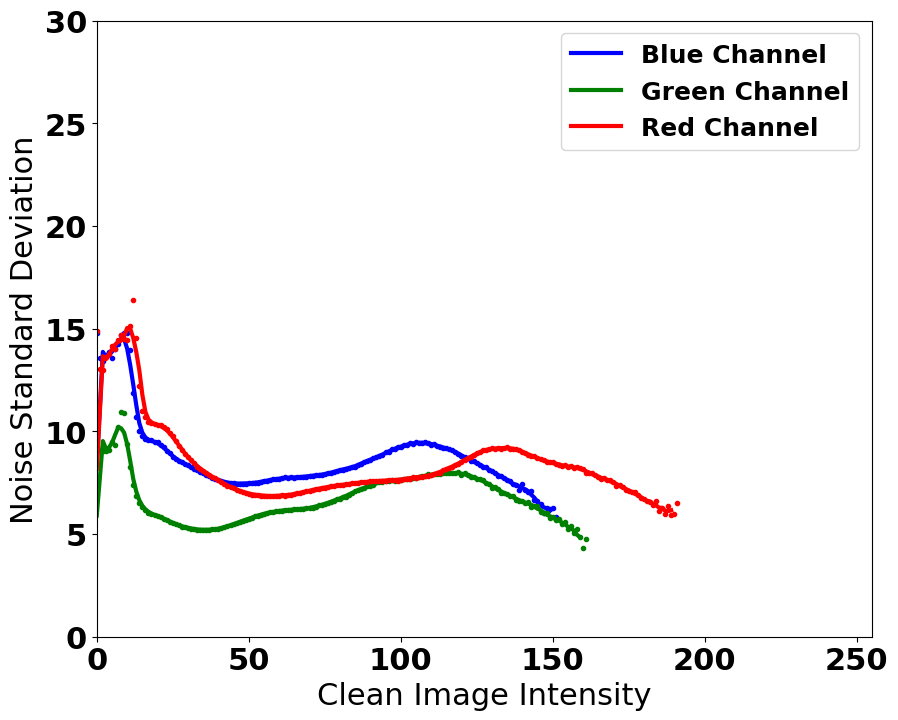}\label{fig:fig_noise_std_vis_a}} 
    \hfill
    \subfloat[Scene: 8, Cam: IP, ISO: 400]
    {\includegraphics[width=\wp]{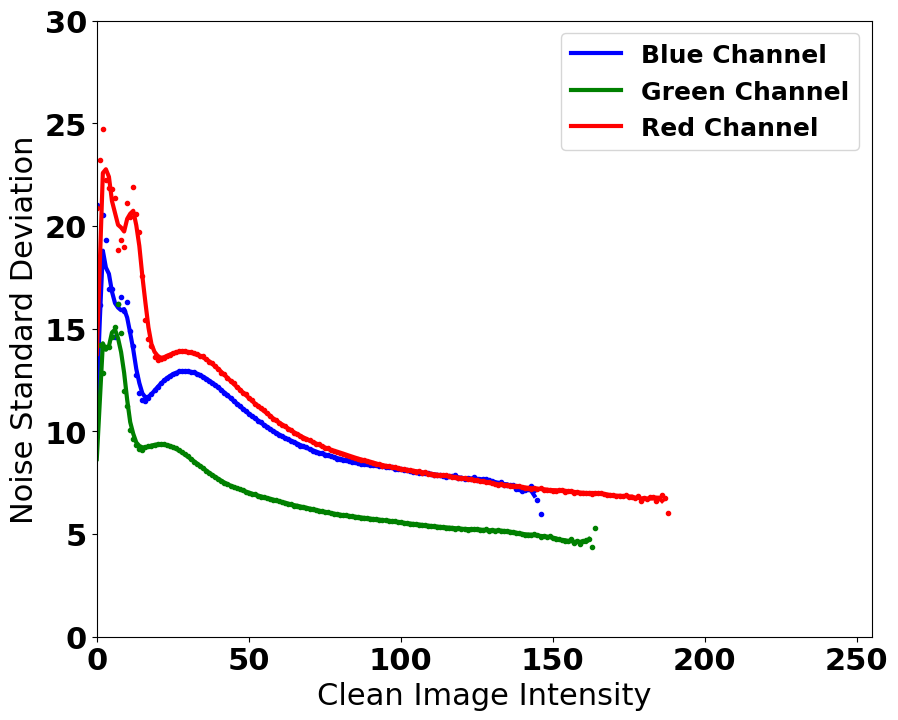}\label{fig:fig_noise_std_vis_b}} 
    \hfill
    \subfloat[Scene: 7, Cam: IP, ISO: 400]
    {\includegraphics[width=\wp]{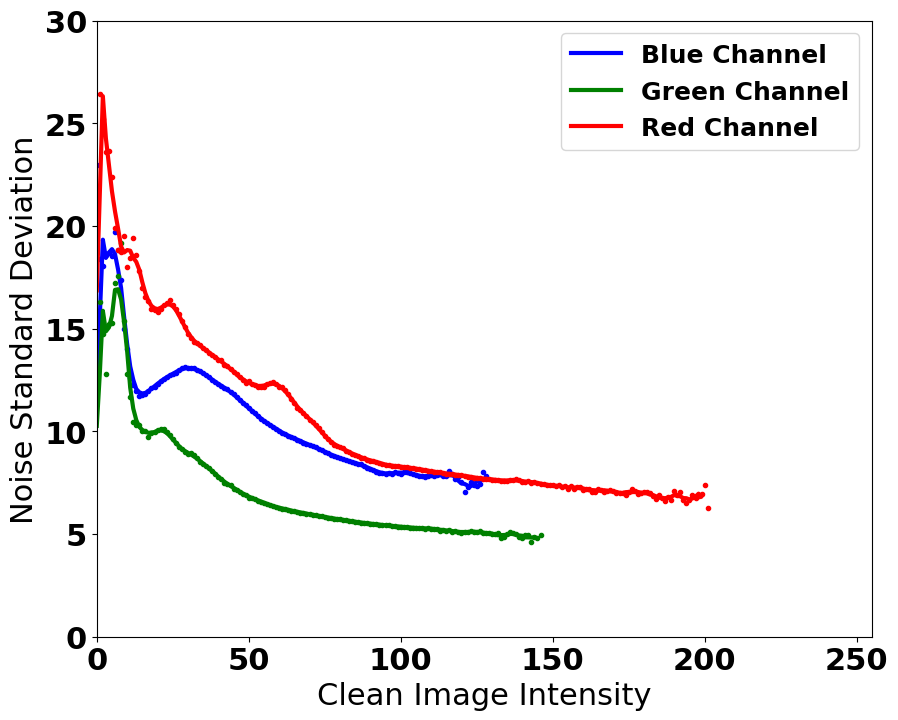}\label{fig:fig_noise_std_vis_c}} 
    \hfill
    \caption{
        \textbf{Illustrations of the relationship between the standard deviation of the sRGB noise and clean image intensity under various conditions.} These plots demonstrate that sRGB noise exhibits unpredictable and complex distributions. In addition, they show that the noise distribution of the sRGB noise varies depending on conditions such as camera type, ISO, and scene. 
        }
   \label{fig:fig_noise_std_vis}
\end{figure*}

\subsection{Normalizing Flows}
\noindent Normalizing Flows is another well-known generative model in computer vision. Compared to GAN, Normalizing Flows shows higher training stability even with small-sized datasets. This feature makes them effective for sRGB noise modeling where the available datasets are often limited in size. Recently, Kousha \textit{et al.}~\cite{srgb_flow} proposed a conditional Normalizing Flows-based noise modeling method in the sRGB image domain. This method allows stable training of a noise model considering clean images and camera conditions. However, due to the constraints of Normalizing Flows where each layer must be invertible, the expressiveness of the transformation function is limited. Consequently, this leads to performance limitations in learning data representations. Therefore, the method has limitations in synthesizing realistic noise images compared to GAN-based models. Specifically, there are issues in modeling the spatial correlation of the noise, which is commonly present in the sRGB image domain.
To address these challenges, Kim \textit{et al.}~\cite{naflow} proposed a method called NAFlow, which focuses on improving the handling of spatially correlated noise through the use of multi-scale noise embedding techniques and Gaussian mixture models. By incorporating these techniques, NAFlow effectively models the spatial correlation of noise. However, NAFlow has a limitation in that it uses real-world noisy images paired with clean images to synthesize the noise. Using real noisy-clean image pairs limits NAFlow's ability to perform noise synthesis in real-world scenarios. This constraint specifically restricts its applicability, such as when processing images captured in uncontrolled environments such as dynamic scenes where obtaining noisy-clean image pairs is not feasible. As a result, the challenge of effectively modeling the spatial correlation of noise using Normalizing Flows remains.
To sum up, while both GAN and Normalizing Flows have been actively adopted for noise modeling in the sRGB image domain, each method has its own strengths and weaknesses. Unlike previous attempts to model noise with a single generative model, we propose a hybrid approach that leverages the strengths of both models. By exploiting GAN's capability in extracting high-dimensional noise features to model spatial correlation and Normalizing Flows' high training stability with small-sized datasets, our approach effectively captures the factors affecting noise distribution, such as camera conditions, without requiring separate neural networks for each camera type. Additionally, our method does not require the paired noisy image corresponding to the clean image for noise synthesis, further enhancing its applicability in diverse real-world scenarios.

\section{Method}
\label{sec:method}

\begin{figure*}[t]
  \centering
    \renewcommand{\wp}{0.40\linewidth}
    \centering
    {\includegraphics[width=\wp]{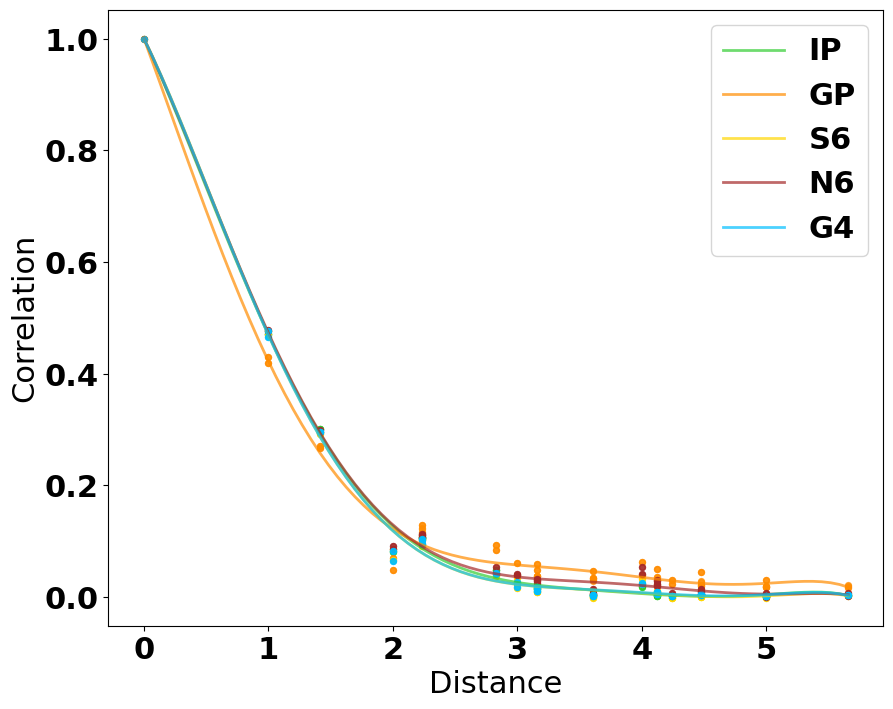}\label{fig:fig_spatial_correlation_a}} 
    {\includegraphics[width=\wp]{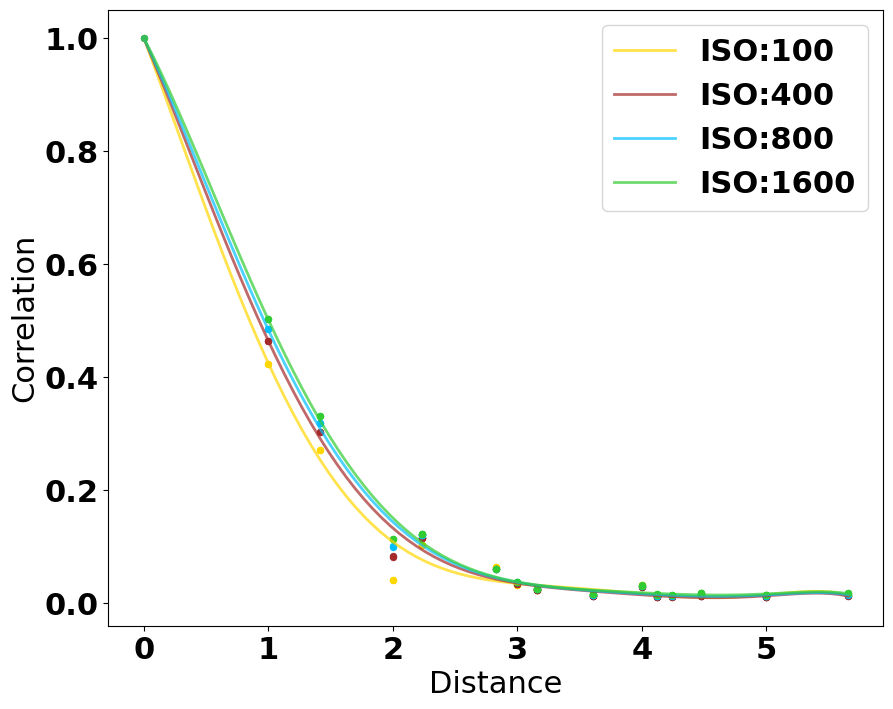}\label{fig:fig_spatial_correlation_b}} 
    \caption{
        \textbf{An illustration of the correlation $r$ versus distance $d$ between neighboring pixels under various camera types and ISO settings.} This figure shows that the noise value of a pixel is highly correlated with its neighboring pixels and shows similar behavior across camera types or ISO settings.
        }
   \label{fig:fig_spatial_correlation}
\end{figure*}

\subsection{Analyzing sRGB Noise}
\noindent sRGB noise has unpredictable and complex distributions. For accurate noise modeling, identifying the factors that affect its distribution is crucial. Therefore, we analyzed the distributions of sRGB noise from the SIDD dataset under different scenarios. Our analysis of the SIDD dataset, summarized in Figure~\ref{fig:fig_noise_std_vis} and \ref{fig:fig_spatial_correlation}, confirms that sRGB noise is affected by three main factors: clean image intensity, image structure, and spatial correlation.
For accurate noise modeling, it is essential to consider all factors that affect the noise. Detailed descriptions of each factor are given below.

The primary factor affecting the sRGB noise distribution is the clean image intensity. We first present the noise distribution in the raw-RGB image domain to clearly demonstrate this relationship. In the raw-RGB domain, one of the most commonly used noise models is the heteroscedastic Gaussian noise~\cite{foi2009clipped}. This model describes raw-RGB noise as a signal-dependent noise with a zero mean and Gaussian distribution. Given a paired raw-RGB clean image $X$ and its noisy counterpart $Y$ such that $Y=X+N$, the Gaussian distribution of noise at pixel $i$ can be represented as: 
\begin{equation}
  N_i \sim \mathcal{N}(0,\beta_s^2 \cdot X_i +\beta_c^2),
  \label{eq:eq1}
\end{equation}
where $\beta_s$, $\beta_c$ are noise distribution parameters that represent signal-dependent and signal-independent. These parameters are determined for each channel and are affected by the camera type and ISO setting.
Given that sRGB images are transformed version of raw-RGB images via the ISP, a clean sRGB image $x$ and noisy sRGB image $y$ can be represented as $y=x+n$. This relationship can be related to their paired raw-RGB images by the equation:
\begin{equation}
  (x,y)=(I(X),I(Y)), 
  \label{eq:eq2}
\end{equation}
where $I(\cdot)$ denotes the ISP. From this equation, we can deduce that sRGB noise is affected by the noise present in raw-RGB; therefore, it can be affected by the clean image intensity, camera type, and ISO setting. Figure~\ref{fig:fig_noise_std_vis_a} and \ref{fig:fig_noise_std_vis_b} demonstrate that sRGB noise is affected by the clean image intensity in a complex, non-linear fashion and varies depending on the camera type and ISO setting.
In addition, the distribution of noise in the sRGB domain is also affected by image structures. This is evident when comparing Figure~\ref{fig:fig_noise_std_vis_b} and \ref{fig:fig_noise_std_vis_c}; despite having the same camera type, ISO settings, and illuminant settings, the noise distribution can vary depending on the captured scene. This variability is due to signal processing techniques, such as local adjustments in the ISP. Therefore, to estimate the noise variation of a specific pixel, it is essential to consider the surrounding image structure. 
Considering the two previously mentioned factors - clean image intensity and image structure, the mean and standard deviation of the pixel-wise noise distribution for the channel $c$ of the $i$th pixel can be expressed by the following equation:
\begin{equation}
    (\mu_{i,c},\sigma_{i,c})=f(P(x_{i,c}), \delta, \gamma),
  \label{eq:eq3}
\end{equation}
where $\delta$ and $\gamma$ denote the camera type and ISO settings respectively, $P(\cdot)$ denotes the local image patch composed of the pixel $x_{i,c}$ and its neighboring pixels, and $f(\cdot)$ denotes the non-linear relationship between the parameters of the noise distribution and the factors that affect it. This allows us to model the pixel-wise noise $v$ at each pixel as a Gaussian distribution:
\begin{equation}
    v_{i,c}\sim  \mathcal{N}(\mu_{i,c},\sigma_{i,c}^2).
  \label{eq:eq4}
\end{equation}
The final factor to consider in the distribution of sRGB noise is spatial correlation. Previously, we introduced pixel-wise noise, assuming that the distribution of sRGB noise is independent for each pixel. However, as shown in Figure~\ref{fig:fig_spatial_correlation}, the actual sRGB noise of each pixel is highly correlated with the noise values of its neighboring pixels. This spatial correlation arises due to the demosaicing process of the ISP on the Bayer filter~\cite{bayer1,bayer2}, which interpolates from neighboring noisy subpixels. Therefore, it is crucial to account for this correlation during noise synthesis.

Spatial correlation arises mainly due to the Bayer demosaicing process, which is typically performed without regard to camera settings~\cite{menon, malvar, apbsn}. As shown in Figure~\ref{fig:fig_spatial_correlation}, spatial correlation is consistent regardless of camera type or ISO settings, unlike pixel-wise noise distribution, which is affected by a multitude of factors.

To analyze this inter-pixel relationship shown in Figure~\ref{fig:fig_spatial_correlation}, we used Pearson correlation. The correlation $r$ is calculated as:
\begin{equation}
r = Cov(N,\tilde{N})/(\sigma_N \cdot \sigma_{\tilde{N}}),
\label{eq:eq_correlation_pearson}
\end{equation}
where $N$ represents the set of pixels in the noise image, and $\tilde{N}$ denotes the set of pixels at a specific distance from each pixel:
\begin{equation}
\tilde{N}(x,y) = N(x+d_x,y+d_y),
\label{eq:eq_correlation_neighbor}
\end{equation}
with the distance $d$ between pixels calculated as:
\begin{equation}
d=\sqrt{d_x^2 + d_y^2}.
\label{eq:eq_correlation_distance}
\end{equation}

Considering spatial correlation, the sRGB noise for each pixel can be modeled using the following equation:
\begin{equation}
n_{i,c}=g(P(v_{i,c})),
\label{eq:eq_correlation}
\end{equation}
where $g(\cdot)$ denotes a function that maps spatial correlation between the pixel $v_{i,c}$ of the noise image and its neighboring pixels.

\subsection{Hybrid Architecture}
\begin{figure*}[t]
  \centering
    \includegraphics[width=0.9\linewidth]{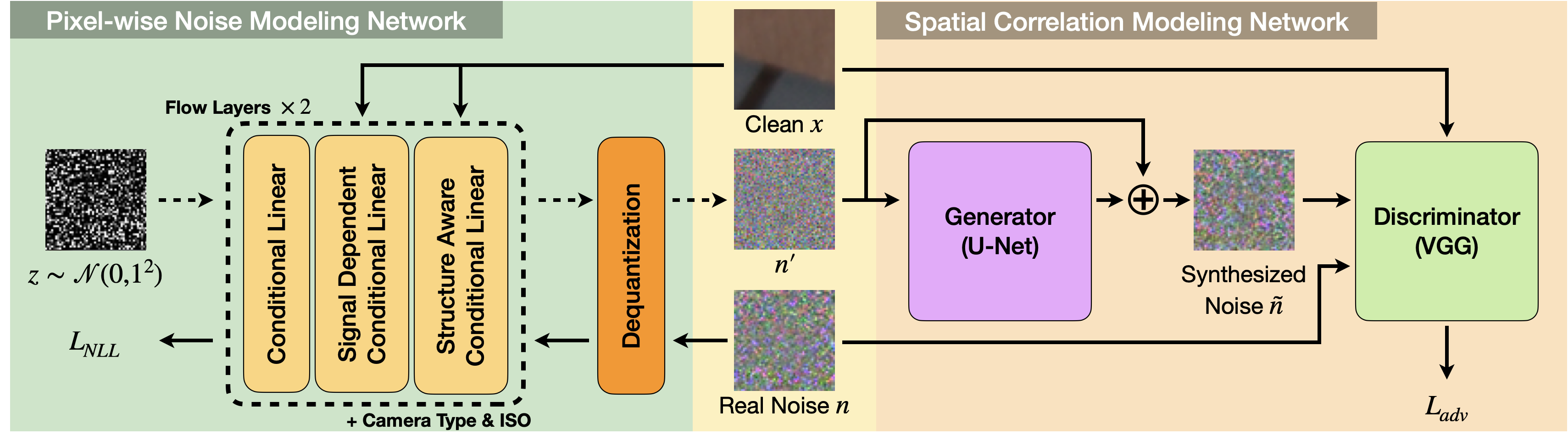}
    \caption{
        \textbf{The overall architecture of our proposed framework for noise synthesis.} Our NM-FlowGAN is comprised of two main components: the pixel-wise noise modeling network and the spatial correlation modeling network. These networks are based on Normalizing Flows and GAN, respectively. We employ a dequantization layer at the beginning of the pixel-wise noise modeling network, which adds uniformly sampled noise to the images during training. For better visualization, the magnitudes of the noise images are amplified. 
        }
    \label{fig:fig_overall_architecture}
\end{figure*}

\noindent Inspired by our analysis, we introduce our novel sRGB noise modeling method named NM-FlowGAN, which is a hybrid approach that combines Normalizing Flows and GAN. As shown in Figure~\ref{fig:fig_overall_architecture}, our proposed neural network for synthesizing real sRGB noise includes two main components: the pixel-wise noise modeling network and the spatial correlation modeling network. 

The pixel-wise noise modeling network based on Normalizing Flows assumes that the distribution of noise value for each pixel is identical. As mentioned above, pixel-wise noise is affected not only by the clean image intensity and image structures but also by camera conditions such as camera type and ISO settings. Considering these aspects, a network based on Normalizing Flows, which ensures stable training even with small-sized datasets, is advantageous for pixel-wise noise modeling. However, networks based on Normalizing Flows are not well-suited for modeling spatial correlation.
To model spatial correlation, it is crucial to consider the complex relationships among neighboring pixels. However, due to the invertibility constraint in normalizing flows, it is challenging to fully capture these relationships. For example, affine coupling~\cite{real_nvp} is adopted in sRGBFlow~\cite{srgb_flow} to model spatial correlation, which considers only a subset of input channels to estimate the scale and translation factors applied to the remaining channels. The limitation of not being able to generate an output feature that takes into account all input features is a constraint on representing complex correlations between neighboring pixels. Furthermore, sRGBFlows primarily focus on learning representations through inter-channel operations such as affine coupling and invertible $1\times1$ convolution~\cite{glow}, they are less suited for modeling pixel-to-pixel spatial correlations.
To address this issue, we adopt a GAN-based spatial correlation modeling network. GAN-based methods are well-suited for handling high-dimensional and complex data distributions. In addition, unlike Normalizing Flows, there are no constraints on designing network architecture, allowing for flexible modeling of pixel-to-pixel relationships. As mentioned above, GAN-based models have a weakness of low training stability on small-sized datasets. However, spatial correlation is not affected by camera type or ISO setting. As a result, this shortcoming is not evident when modeling spatial correlations.
%

\subsection{Pixel-wise Noise Modeling Network}

\noindent In our pixel-wise noise modeling network, the distribution of each pixel is assumed to be identical, and Gaussian distributions are estimated for each pixel. Our network, based on Normalizing Flows, comprises three flow layers: conditional linear flow, signal-dependent conditional linear flow (SDL), and structure-aware conditional linear flow (SAL). Each layer effectively estimates the pixel-wise noise distribution by considering the factors affecting it.

\begin{figure*}[thb!]
  \centering
    \includegraphics[width=0.9\linewidth]{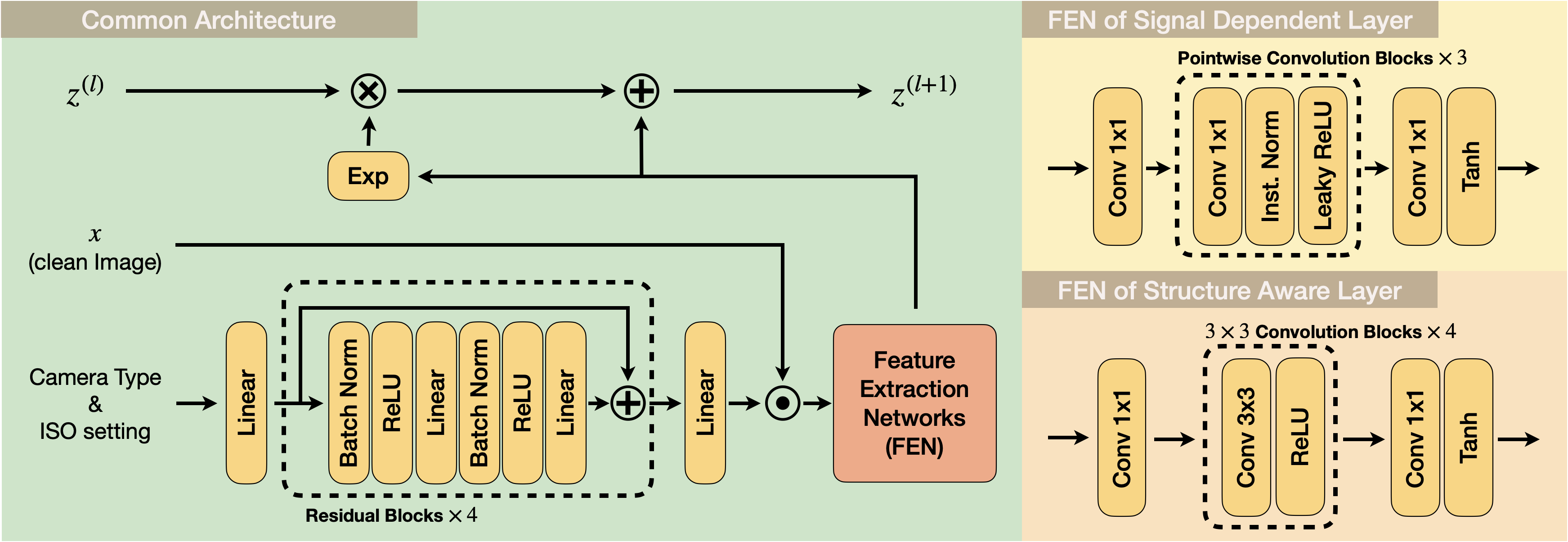}
    \caption{
        \textbf{The detailed architectures of our conditional linear flow layers: SDL and SAL.} These layers are invertible, but we describe the forward pass of conditional linear flows. In the figure, $\odot$ denotes channel-wise concatenation, $\oplus$ and $\otimes$ denote element-wise addition and multiplication, respectively.
        }
   \label{fig:fig_detailed_architecture}
\end{figure*}

Recently, a simple invertible form of the conditional linear flow layer has been proposed in sRGBFlow~\cite{srgb_flow}. This layer linearly transforms the input feature using values determined by the camera type $\delta$ and ISO setting $\gamma$:
\begin{equation}
    z^{(l+1)}_i=z^{(l)}_i\otimes F^{(l)}_{cs}(\delta,\gamma)+F^{(l)}_{ct}(\delta,\gamma),
\label{eq:eq6}
\end{equation}
where $z^{(l)}$ denotes the input of the $i$th pixel in the $l$th flow layer, $\otimes$ denotes element-wise multiplication, and $F_{cs}(\cdot)$ and $F_{ct}(\cdot)$ are simple neural networks designed to output the scale and translation factors, considering the values of $\delta$ and $\gamma$. While this layer can perform a transformation that considers the camera conditions, the non-linearity and complexity of real sRGB noise make it difficult to model accurately with such a simple transformation.

To address these limitations, we designed enhanced layers that estimate the pixel-wise noise distribution by considering factors such as the clean image intensity and the image structure. SDL performs a linear transformation based on the clean image intensity:
\begin{equation}
    z^{(l+1)}_i = z^{(l)}_i \otimes F^{(l)}_{ss}(x_i, \delta, \gamma) + F^{(l)}_{st}(x_i, \delta, \gamma),
\label{eq:eq7}
\end{equation}
where $F_{ss}(\cdot)$ and $F_{st}(\cdot)$ consist of pointwise convolution layers and non-linear activation layers, designed to output scale and translation factors based on the clean image intensity and camera conditions. SAL performs a linear transformation by considering the surrounding structure of pixel $i$:
\begin{equation}
    z^{(l+1)}_i = z^{(l)}_i \otimes F^{(l)}_{ts}(P(x_i), \delta, \gamma) + F^{(l)}_{tt}(P(x_i), \delta, \gamma),
\label{eq:eq8}
\end{equation}
where $F_{ts}(\cdot)$ and $F_{tt}(\cdot)$ are designed to have limited receptive fields with $3\times3$ convolution layers and non-linear activation layers. They output scale and translation factors considering both the image structure and camera conditions.

Figure~\ref{fig:fig_detailed_architecture} illustrates the detailed architecture of the core components of our pixel-wise noise modeling network: SDL and SAL. These layers are designed to estimate and apply scale and bias values for each pixel using a clean image and conditions, including camera type and ISO setting. SDL and SAL first apply a one-hot encoding of the camera type and the ISO setting, and then extract the features using an architecture consisting of residual blocks. This method of encoding camera conditions is adopted from sRGBFlow~\cite{srgb_flow}. The extracted features are then channel-wise concatenated with the clean image to serve as input to the feature extraction networks (FEN) of SDL and SAL. Each of these layers has a different architecture for feature extraction, and the outputs from these layers are used as scale and bias parameters for each pixel.

Specifically, SDL consists mainly of pointwise convolution layers, instance normalization layers, and activation layers. This design aims to estimate log scale and bias parameters based on the clean image intensity of each pixel. On the other hand, SAL consists mainly of convolution layers with $3\times3$ filters and activation layers. Its design aims to estimate log scale and bias parameters based on the image structure surrounding each pixel.

Like other architectures that rely on Normalizing Flows, our pixel-wise noise modeling network is trained to minimize the negative log-likelihood (NLL) of the data. Once trained, we can employ this network to synthesize pixel-wise noise $n'$. This synthesis is achieved by sampling from the base distribution $z \sim p_z$ and subsequently applying the inverse flow:
\begin{equation}
    n' = F^{-1}(z | x, \delta, \gamma),
\label{eq:eq9}
\end{equation}
where $F(\cdot)$ denotes our invertible pixel-wise noise modeling network.

\begin{figure}[t!]
  \centering
    \includegraphics[width=0.8\linewidth]{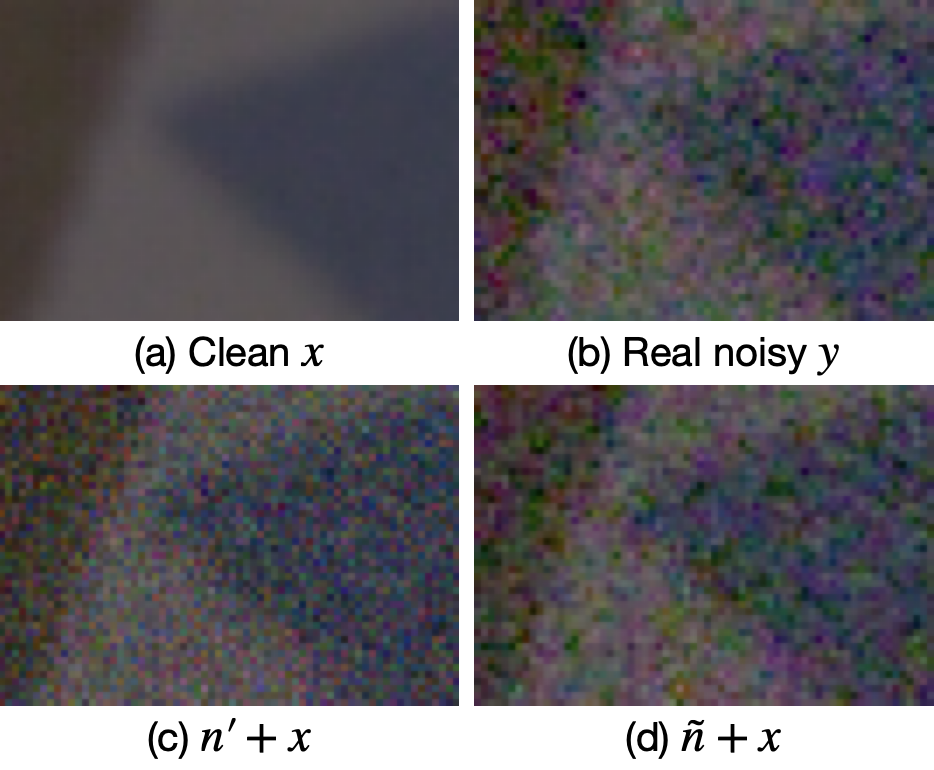}
    \caption{
        \textbf{Visualization of sRGB noisy image synthesis.} (a) clean image $x$, (b) real noisy image $y$, (c) synthesized pixel-wise noisy image $n'+x$, and (d) final synthesized noisy image considering spatial correlation $\tilde{n}+x$.
        }
   \label{fig:fig_pixelwise_noise}

\end{figure}

\subsection{Spatial Correlation Modeling Network}

\noindent Given the synthesized pixel-wise noise, our GAN-based spatial correlation modeling network models the pixel-to-pixel correlation to reduce the discrepancy between the real noise $n$ and synthesized pixel-wise noise $n'$. Specifically, as represented in \eqref{eq:eq_correlation}, the final noise value is determined by considering the surrounding pixel-wise noise values. We adopt the U-Net~\cite{unet} for the generator network and a VGG network-based architecture~\cite{vgg} for the discriminator network. Consequently, by using our spatial correlation modeling network $G(\cdot)$, we can generate the final synthesized sRGB noise image $\tilde{n}$ as follows:
\begin{equation}
    \tilde{n} = G(n') + n'.
\label{eq:eq10}
\end{equation}

Figure~\ref{fig:fig_pixelwise_noise} provides visual examples of a pixel-wise noisy image and a noisy image considering spatial correlation.

\begin{table*}[t]
  \centering
  \renewcommand\arraystretch{1.2}
  \setlength{\tabcolsep}{14pt}

  \begin{center}
    \caption{
    \textbf{KL divergence between synthesized noise and real noise calculated under various camera types.} $\dagger$ denotes the method of synthesizing sRGB noise with a paired noisy image. In addition, G4, GP, IP, S6, and N6 represent the average KL divergence for the respective camera types, and AVG represents the average KL divergence of the entire test set.
    }\label{table:table_kld_baselines_wide}
    
    \vspace{0mm}
    \scalebox{1.0}{
    \begin{tabular}{cccccccccc}
      \toprule
      & \multirow{1}{*}{\begin{tabular}[c]{@{}c@{}}\textbf{Method} \end{tabular}} 
      & \textbf{G4}
      & \textbf{GP} 
      & \textbf{IP}
      & \textbf{N6}
      & \textbf{S6}
      & \textbf{AVG$^\downarrow$}\\

      \midrule
      \multirow{2}{*}{\begin{tabular}[c]{@{}c@{}} \textbf{Traditional Methods} \end{tabular}}
      & AWGN & 0.0672 & 0.1136 & 0.0559 & 0.1467 & 0.0836	 & 0.0856        \\
      & Heteroscedastic & 0.0718 & 0.1393 & 0.0575 & 0.1330 & 0.0823	 & 0.0779 	    \\
      
      \midrule
      \multirow{2}{*}{\begin{tabular}[c]{@{}c@{}} \textbf{Learning-based Methods} \\ (with paired image) \end{tabular}}
      & NeCA-W$^\dagger$~\cite{neca} & 0.0611 & 0.0260 & 0.0284 & 0.0435 & 0.0337 & 0.0344 \\
      & NAFlow$^\dagger$~\cite{naflow} & 0.0186 & 0.0681 & 0.0129 & 0.0423 & 0.0164 & 0.0286 \\ 
      \midrule
      \multirow{4}{*}{\begin{tabular}[c]{@{}c@{}} \textbf{Learning-based Methods} \\ (without paired image)\end{tabular}}
      
      & C2N~\cite{c2n} & 0.2296 & 0.4460 & 0.1172 & 0.2399 & 0.6822 & 0.3878 \\
      & NeCA-W~\cite{neca} & 0.0893 & 0.0867 & 0.0452 & 0.0847 & 0.0716 & 0.0706  \\
      & sRGBFlow~\cite{srgb_flow} & 0.0424 & 0.0669 & 0.0390 & 0.0528 & 0.0439 & 0.0477 \\
      & \textbf{Ours} & \textbf{0.0389} & \textbf{0.0259} & \textbf{0.0164} & \textbf{0.0407} & \textbf{0.0334} & \textbf{0.0283} \\

      \bottomrule
    \end{tabular}}

    \end{center}
  \vspace{0mm}

\end{table*}

\subsection{Loss Functions}

{\noindent \bf{NLL Loss:}}
Our pixel-wise noise modeling network $F$ is composed of a sequence of $L$ flow layers: 
\begin{equation}
    F=F^{(0)}\circ F^{(1)}\circ \cdots \circ F^{(L-1)}. 
\label{eq:eq11}
\end{equation}
For each layer, the input and output relations can be expressed as:
\begin{equation}
    z^{(l+1)}=F^{(l)}(z^{(l)}|x,\delta,\gamma).
\label{eq:eq12}
\end{equation}without the necessity of training separate neural networks for each camera or using paired noisy images.
For the sake of clarity, we define $z^{(0)} \stackrel{\text{def}}{=} n
 $ and  $z^{(L)} \stackrel{\text{def}}{=} z$ for conciseness. To train this network, we employ the  NLL loss. It can be formulated as:
\begin{multline}
    L_{NLL}(n|x,\delta,\gamma)=-p_z(F(n|x,\delta,\gamma)) \\
    -\sum_{l=0}^{L-1} log|det(\mathcal{D}F^{(l)}(z^{(l)}|x,\delta,\gamma))|,
\label{eq:eq13}
\end{multline}
where $\mathcal{D}F(\cdot)$ denotes a function that outputs the Jacobian matrix of $F$ at the given input.

{\noindent \bf{GAN Loss:}}
To train our generator network $G$, we employ WGAN-GP~\cite{wgan, wgan-gp} approach to calculate the adversarial loss. The adversarial loss for $G$ is given by:
\begin{equation}
L_{adv}(x,n')=-\lambda \cdot D(x||G(sg(n'))),
\label{eq14}
\end{equation}
where $D(\cdot)$ denotes our discriminator, and $x$ and $n'$ denote the clean image and the synthesized pixel-wise noise image, respectively. In addition, $\lambda$ is the balance parameter between the pixel-wise noise modeling networks and the spatial correlation modeling networks, the operator $||$ denotes the channel-wise concatenation, and $sg(\cdot)$ denotes the stop-gradient operation~\cite{stop-gradient}. This equation indicates that our discriminator evaluates the authenticity of the noise synthesized by $G$, with consideration of the clean image $x$. Meanwhile, to ensure that the discriminator is effectively trained with the generator, we define the critic loss $L_{critic}$ for our discriminator as follows:
\begin{equation}
L_{critic}=\lambda \cdot (L_{wgan}+\alpha L_{gp}),
\label{eq15}
\end{equation}
where $\alpha$ is the weight for the gradient penalty term, set to 10 in all our experiments, and $L_{wgan}$ is the original critic loss, defined as:
\begin{equation}
L_{wgan}(x,n,n') = D(x||G(sg(n'))) - D(x||n).
\label{eq16}
\end{equation}
In addition, $L_{gp}$ is the gradient penalty loss, defined as:
\begin{equation}
L_{gp}(x,n,n')=(||\nabla_{\hat{x}}D(\hat{x})||_2 -1)^2,
\label{eq17}
\end{equation}
where $\hat{x}=\epsilon(x||n)+(1-\epsilon)(x||G(sg(n')))$ and $\epsilon \sim U(0,1)$.

\section{Experiments}
\label{sec:experiments}

\subsection{Experimental Setup}
{\noindent \bf{Dataset:}}
To evaluate the performance of sRGB noise modeling, we use the Smartphone Image Denoising Dataset (SIDD~\cite{sidd}). The SIDD medium split contains 320 noisy-clean image pairs captured with various camera types: LG G4 (G4), Google Pixel (GP), iPhone 7 (IP), Motorola Nexus 6 (N6), and Samsung Galaxy S6 Edge (S6), each under various ISO settings. For a fair comparison, we follow the sRGBFlow~\cite{srgb_flow} dataset settings, using 80\% of the data for training and 20\% of the data for validation.

{\noindent \bf{Implementation Details:}}
We extract patches of size $96\times96$ with a step size of $48$ to optimize NM-FlowGAN\footnote{Our code is available at: \url{https://github.com/YoungJooHan/NM-FlowGAN}}. During training, we augment all of the training patches by randomly flipping and rotating them by $90^{\circ}$. We employ the Adam~\cite{ADAM} optimizer for training with an initial learning rate set to $10^{-4}$. The learning rate decreases by half every 10 epochs, and our model trains over 40 epochs. In addition, we set $\lambda$ to $0.5$ in our experiments. 

\begin{figure*}[t]
  \centering
    \includegraphics[width=0.9\linewidth]{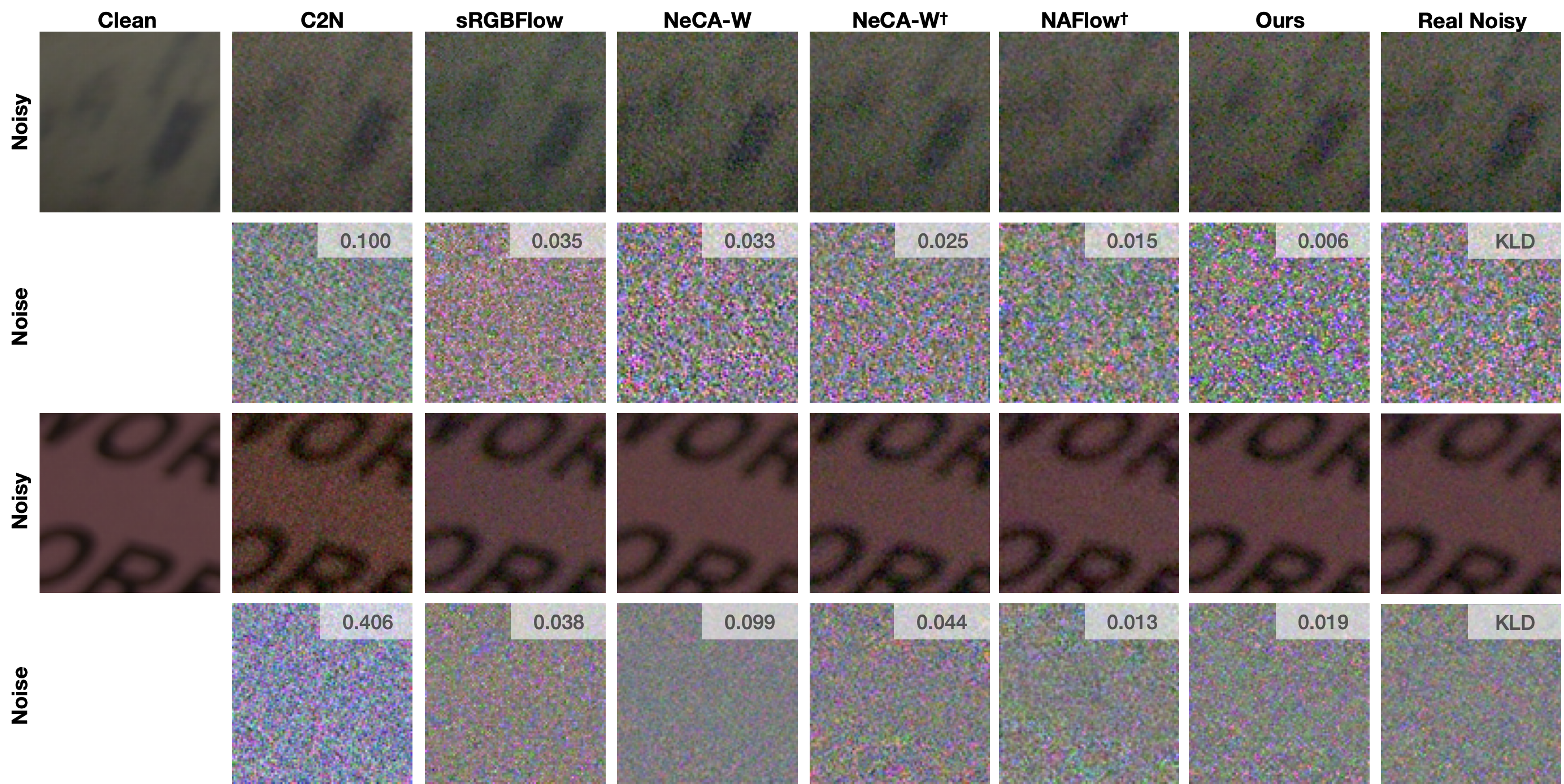}
    \caption{
        \textbf{Visualization of synthetic noisy and noise images from our NM-FlowGAN and other baselines.} The images in the second (fourth) row represent the noise of the images in the first (third) row. For better visualization, the magnitudes of the noise images are amplified. Best view in zoom-in.
        }
   \label{fig:fig_synthetic_noise}

\end{figure*}

{\noindent \bf{Baselines:}}
Our NM-FlowGAN is compared with several existing sRGB noise modeling methods in the experiments. We include traditional methods such as Additive White Gaussian Noise(AWGN), heteroscedastic Gaussian noise, and deep learning-based modeling methods such as C2N~\cite{c2n}, sRGBFlow~\cite{srgb_flow}, NeCA-W~\cite{neca}, and NAFlow~\cite{naflow}. For the synthesis using AWGN and heteroscedastic Gaussian noise, we estimate the parameters corresponding to each noise model and apply them to each pixel. For C2N, sRGBFlow, NeCA-W, and NAFlow, we use the official codes provided by the authors. Specifically, $\dagger$ in Table~\ref{table:table_kld_baselines_wide} and Figure~\ref{fig:fig_synthetic_noise} refers to a method that uses paired noisy images for noise synthesis.

Using paired noisy images allows for more accurate noise generation. However, the need for paired noisy images for noise synthesis does not take one of the main advantages of noise modeling mentioned in Section 1, which is the ability to generate noisy-clean image pairs in scenes that are not achievable in real-world environments. Therefore, its applicability in real-world scenarios is limited. For NeCA-W, since the method can perform noise synthesis using both paired noisy images and unpaired noisy images, we present the experimental results for both scenarios.

\subsection{Results on sRGB Noise Modeling}
\noindent Table~\ref{table:table_kld_baselines_wide} lists the discrete Kullback-Leibler(KL) divergence results for NM-FlowGAN and several baseline models. We evaluate the performance of the noise modeling using the KL divergence, which determines the similarity between the distributions of the real and synthesized noise by comparing the histogram. Specifically, the histogram range is set from -260 to 260, with 130 intervals. 
In our experimental results, our NM-FlowGAN showed the best performance among the methods that do not use paired noisy images for noise synthesis. Specifically, NM-FlowGAN shows a lower KL divergence of 0.043 and 0.020 compared to the recently proposed NeCA-W and sRGBFlow, respectively. Notably, our NM-FlowGAN achieves a slightly lower KL divergence compared to NeCA-W$^\dagger$ and NAFlow$^\dagger$, both of which use paired noisy images. Figure~\ref{fig:fig_synthetic_noise} shows a qualitative comparison between the real noise images and the synthesized noise images generated by our NM-FlowGAN and other baselines.

\begin{figure}[t!]
  \centering
    \includegraphics[width=1.0\linewidth]{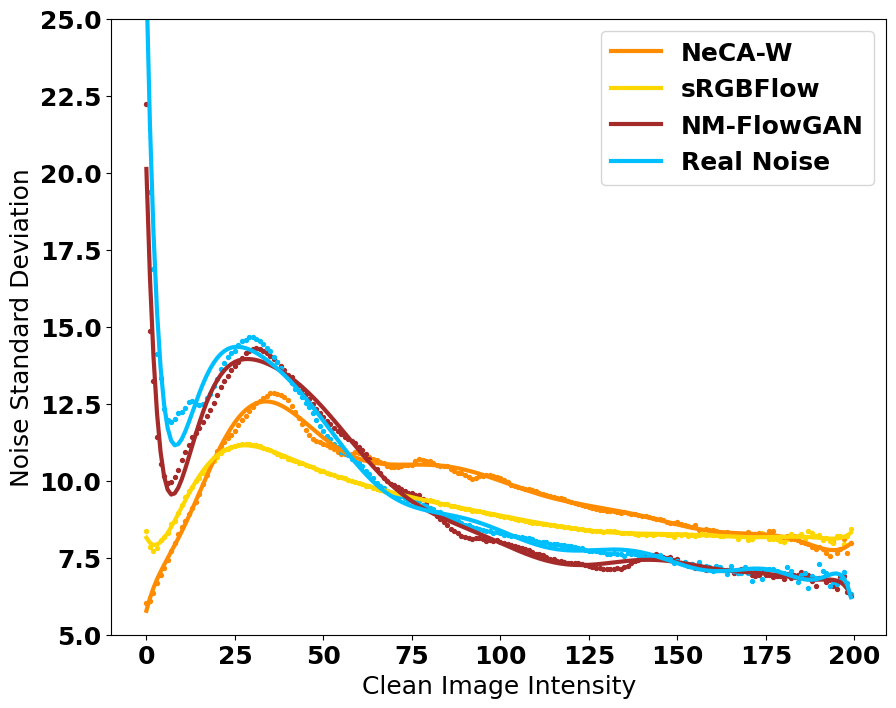}
    \caption{
        \textbf{An illustration of the relationship between the standard deviation of the sRGB noise and the clean image intensity.} This is from a IP camera type, with an ISO setting of 400 in the red channel.
        }
   \label{fig:fig_noise_std_per_methods}
\end{figure}

To further validate the noise modeling performance of our model, we present additional experimental results in this section. First, Figure~\ref{fig:fig_noise_std_per_methods} shows the standard deviation as a function of clean image intensity for our NM-FlowGAN, sRGBFlow, and NeCA-W. This figure indicates that our model shows similar behavior to real noise compared to the other baselines.

\subsection{Ablation Study}
\noindent Table~\ref{table:table_kld_ablation} summarizes the KL divergence for different architectural choices of our proposed NM-FlowGAN. In the experiment using only a GAN-based architecture, we observe a high KL divergence due to the lack of consideration for features that affect the noise distribution.
In the experiments using the flow layers, the KL divergences are gradually decreased with the addition of signal-dependent and structure-aware conditional linear flow layers. Notably, in the experiment consisting only of flow layers (without GAN), it is observed that the KL divergence is marginally higher than when both flow layers and GAN are used simultaneously. This is because KL divergence, which is based on histogram similarity, may not effectively capture factors that are not revealed in the histogram, such as spatial correlation. Therefore, in the following subsection, we further evaluate the closeness of the synthesized noise to the real noise by comparing the performance of image denoising networks trained with the synthesized noise from each method.

\begin{table}[!t]
\centering
\caption{
\textbf{Ablation study on different architectural choices.} \textbf{SAL} the structure-aware conditional linear flow layer, \textbf{SDL} the signal-dependent conditional linear flow layer, and \textbf{GAN} the spatial correlation network based on GAN. 
}
\renewcommand{\arraystretch}{1.3}
\setlength{\tabcolsep}{9pt}
\scalebox{1.0}{
\begin{tabular}{ ccc | c }
\hline
\bf \textsc{SAL} & \bf \textsc{SDL} & \bf \textsc{GAN} & \bf \textsc{KL-Divergence} \\
\hline
\checkmark    & \checkmark    & 	             & 0.0377        \\ 
              &               & \checkmark       & 0.5546	       \\
              & \checkmark    & \checkmark       & 0.1276        \\
\checkmark    &               & \checkmark       & 0.0371        \\
\checkmark    & \checkmark    & \checkmark	     & \textbf{0.0283}         \\
\hline
\end{tabular}
}
\label{table:table_kld_ablation}
\end{table}

\begin{table}[t!]
  \centering
  \renewcommand\arraystretch{1.2}

  \begin{center}
    \caption{
    \textbf{Image denoising results on the SIDD benchmark dataset.} Each denoiser is trained on synthesized noisy-clean image pairs generated by each model. $^\dagger$ denotes the method of synthesizing sRGB noise with a paired noisy image. 
    }\label{table:table_psnr_ssim_baselines_wide}
    
    \vspace{0mm}
    \scalebox{1.0}{
    \begin{tabular}{cccc}
      \toprule
      & \multirow{1}{*}{\begin{tabular}[c]{@{}c@{}}\textbf{Method} \end{tabular}} 
      & \textbf{PSNR$^\uparrow$(dB)} & \textbf{SSIM$^\uparrow$} \\

      \midrule
      \multirow{1}{*}{\begin{tabular}[c]{@{}c@{}} \textbf{Real Noisy} \end{tabular}}
      & Supervised & 37.73 & 0.941        \\
      
      \midrule
      \multirow{2}{*}{\begin{tabular}[c]{@{}c@{}} \textbf{Traditional} \end{tabular}}
      & Heteroscedastic & 32.13 & 0.843    \\
      & AWGN & 32.48 & 0.854        \\
      
      \midrule
      \multirow{2}{*}{\begin{tabular}[c]{@{}c@{}} \textbf{Learning-based} \\ (with paired image) \end{tabular}}
      & NeCA-W$^\dagger$~\cite{neca} & 36.82 & 0.932  \\
      & NAFlow$^\dagger$~\cite{naflow} & 37.22 & 0.935  \\ 
      \midrule
      \multirow{4}{*}{\begin{tabular}[c]{@{}c@{}} \textbf{Learning-based} \\ (without paired image)\end{tabular}}
      
      & C2N~\cite{c2n} & 33.76 & 0.901  \\
      & sRGBFlow~\cite{srgb_flow} & 34.74 & 0.912  \\
      & NeCA-W~\cite{neca} & 35.93 & 0.926   \\
      & \textbf{Ours} & \textbf{37.04} & \textbf{0.932}  \\

      \bottomrule
    \end{tabular}}

    \end{center}
  \vspace{0mm}

\end{table}

\subsection{Results on Real-world Image Denoising}

{\noindent \bf{Preparation:}}
To further investigate the quality of the synthesized noise, we evaluate its performance in real-world image denoising. We use the widely adopted image denoiser DnCNN~\cite{dncnn} to train on the synthesized noisy-clean image pairs generated by each of the baselines. The clean images used to generate the synthesized noisy-clean image pairs are obtained from the sRGB images in the SIDD medium split. In addition, we adopt the SIDD validation and benchmark dataset for validation and testing of image denoising networks. The SIDD validation and benchmark dataset contain 1,280 patches, each of size $256\times256$. Although the ground truth for the SIDD benchmark dataset used in testing is not publicly available, peak signal-to-noise ratio (PSNR) and structural similarity index measure (SSIM) results for denoising can be obtained through the online submission system provided on the SIDD website\footnote{\url{https://www.eecs.yorku.ca/~kamel/sidd/benchmark.php}}. In our real-world image denoising experiments, we include not only the baselines used in sRGB noise modeling but also a supervised method trained on real noisy-clean image pairs to estimate the upper bound performance.

\begin{figure*}[t]
  \centering
    \includegraphics[width=0.8\linewidth]{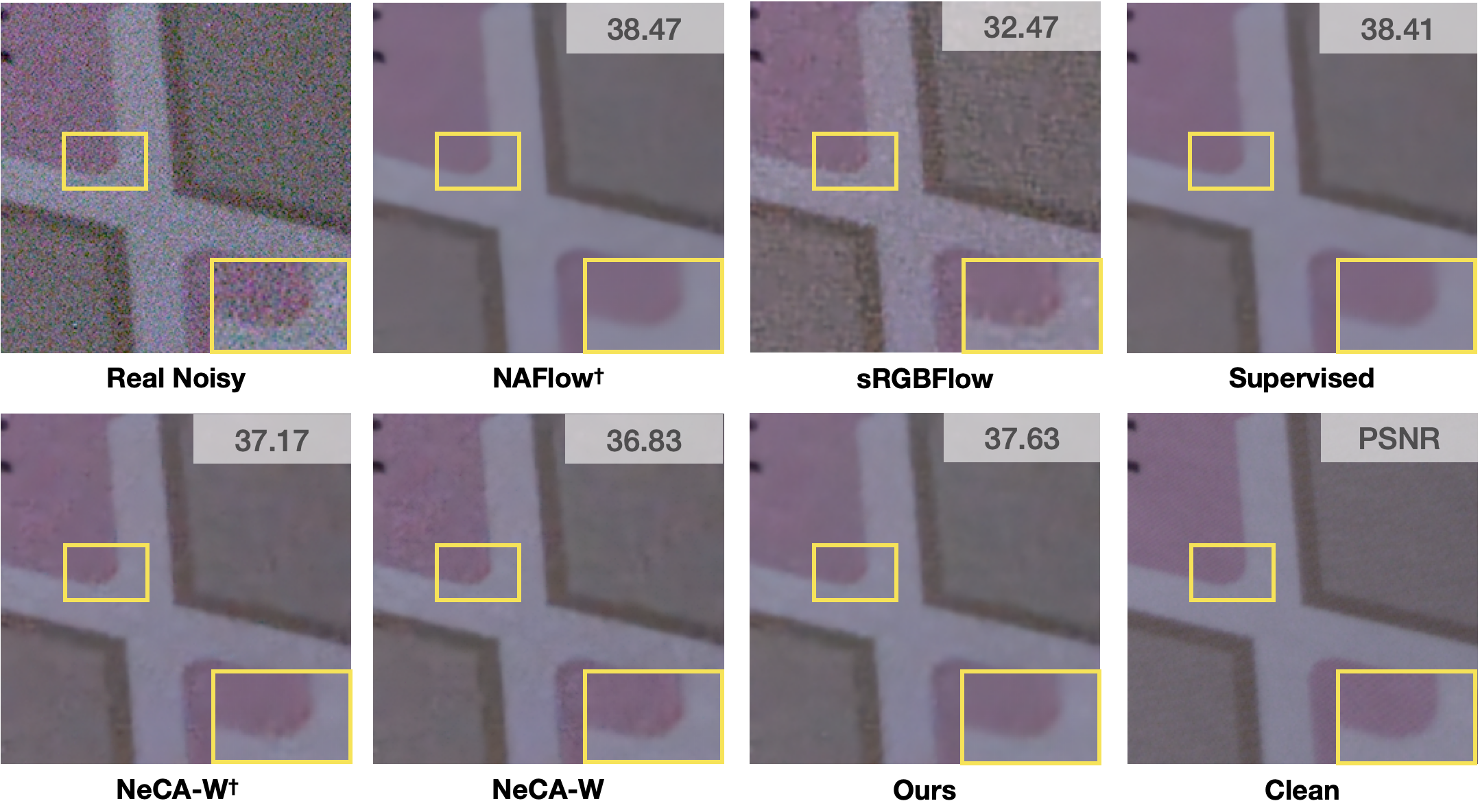}
    \caption{
        \textbf{Visual comparison of denoised images in the SIDD validation dataset.} Each denoiser is trained on synthesized noisy-clean image pairs generated by each model. Best view in zoom-in.
        }
   \label{fig:fig_denoising_baselines}

\end{figure*}
\begin{table}[!t]
\centering
\caption{
\textbf{SIDD validation results under different architectural choices.} The meanings of SAL, SDL, and GAN are as described in Table~\ref{table:table_kld_ablation}.
}
\renewcommand{\arraystretch}{1.3}
\setlength{\tabcolsep}{9pt}
\scalebox{1.0}{
\begin{tabular}{ ccc | cc }
\hline
\bf \textbf{SAL} & \bf \textbf{SDL} & \bf \textbf{GAN} & \bf \textbf{PSNR}(dB) & \bf \textbf{SSIM} \\
\hline
\checkmark    & \checkmark    & 	             & 29.82 & 0.614        \\ 
              &               & \checkmark       & 26.16 & 0.432	       \\
              & \checkmark    & \checkmark       & 36.19 & 0.879        \\
\checkmark    &               & \checkmark       & 36.85 & 0.887        \\
\checkmark    & \checkmark    & \checkmark	     & \textbf{37.07} & \textbf{0.891}        \\
\hline
\end{tabular}
}

\label{table:table_psnr_ssim_ablation}
\end{table}

{\noindent \bf{Results:}}
Table~\ref{table:table_psnr_ssim_baselines_wide} lists the performance of image denoising networks trained on noisy-clean image pairs synthesized by our proposed method in comparison to other baselines. The results indicate that our method outperforms the other baselines in terms of PSNR and SSIM. Specifically, NM-FlowGAN achieves significant PSNR gains of 1.11dB and 2.30dB over NeCA-W and sRGBFlow respectively. These results indicate that the noise generated by our model has a more similar distribution to the real noise compared to the other baselines. A visual comparison between NM-FlowGAN and the baselines is shown in Figure~\ref{fig:fig_denoising_baselines}.
{\noindent \bf{Ablation Study:}}
Table~\ref{table:table_psnr_ssim_ablation} summarizes the image denoising performance of different architecture choices for our method. Notably, the experiment consisting only of flow layers (without GAN), while showing a low KL divergence in Table~\ref{table:table_kld_ablation}, shows a degraded image denoising performance. This indicates that efficient modeling of spatial correlation plays a crucial role in improving the applicability of the noise model. In experiments using the flow layers based on the hybrid architecture (with GAN), both PSNR and SSIM are progressively improved with the addition of signal-dependent and structure-aware conditional linear flow layers.

\begin{table}[!t]
    \caption{\textbf{The effect of enlarging the SIDD dataset with synthesized samples from the SIDD+ dataset~\cite{sidd_plus}.} $+$ denotes that the training dataset has been enlarged by synthesized noisy-clean image pairs generated using the sRGB noise modeling method, and "Supervised" refers to training on the SIDD dataset only.}
  \renewcommand\arraystretch{1.25}
  \centering
  \label{table:table_psnr_ssim_siddplus}
  \setlength{\tabcolsep}{9pt}
    \scalebox{1.0}{
    \begin{tabular}{c | cc}
      \toprule
      \textbf{Method} & \textbf{PSNR(dB)} & \textbf{SSIM} \\
      \midrule
      Supervised & 35.21 & 0.902 \\
      + sRGBFlow~\cite{srgb_flow} & 34.69 & 0.893 \\
      + NeCA-W~\cite{neca} & 35.57 & 0.906 \\
      + \textbf{Ours} & \textbf{35.75} & \textbf{0.908} \\
      \bottomrule
    \end{tabular}}
\end{table}

\subsection{Applicability in Real-world Scenario}
\noindent As mentioned in Section 1, modeling sRGB noise enables the generation of noisy-clean image pairs across various environments, even in scenarios where obtaining such pairs is challenging due to physical constraints. This allows for training image denoising networks with synthesized pairs that are similar to real usage scenarios, thereby enhancing the generalization ability of the model and ensuring robust performance.

\begin{figure}[t!]
  \centering
    \includegraphics[width=0.95\linewidth]{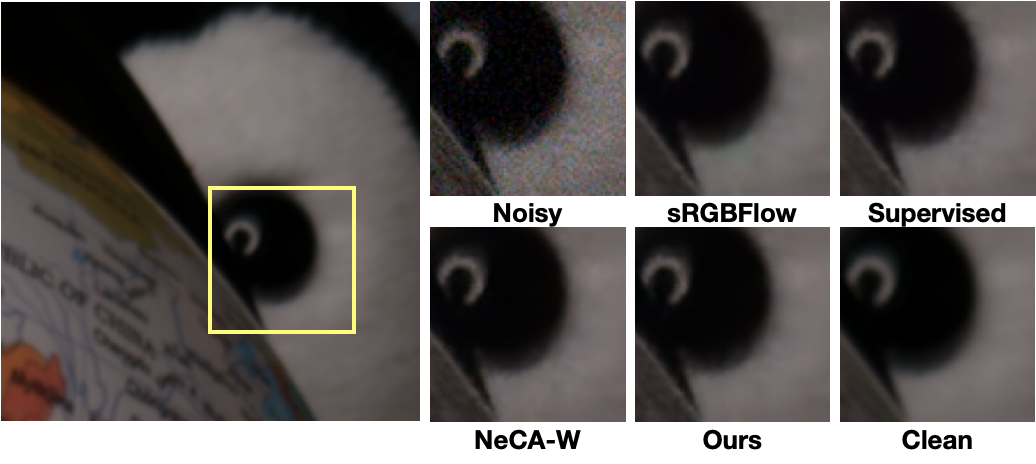}
    \caption{
        \textbf{Visual comparison of denoising results on the SIDD+ dataset.} Each denoiser is trained on the SIDD dataset enlarged by synthesized noisy-clean image pairs generated using the sRGB noise modeling method, and "Supervised" refers to training on the SIDD dataset only.
        }
   \label{fig:fig_siddplus}

\end{figure}

To demonstrate the importance of training with scenes similar to real usage environments, we utilize the SIDD+~\cite{sidd_plus} dataset, which contains scenes different from those in the SIDD dataset. Specifically, we use our NM-FlowGAN to generate synthetic noisy-clean image pairs from 50\% of the SIDD+ dataset, and then add these to the SIDD dataset to train the image denoising networks. Subsequently, we evaluate the performance of the denoiser using the remaining 50\% of the SIDD+ dataset. In the experiment, the camera settings for noise synthesizing in the SIDD+ dataset are randomly selected for each image.
Table~\ref{table:table_psnr_ssim_siddplus} and Figure~\ref{fig:fig_siddplus} demonstrate that image denoising performance improves when networks are trained on datasets that are enlarged with synthesized noisy-clean image pairs derived from scenes similar to actual usage environments. Specifically, when compared to the results using only the SIDD dataset, enlarging the dataset with synthesized noisy-clean image pairs generated by our NM-FlowGAN results in a PSNR gain of 0.54. Furthermore, our proposed method achieves higher gains in PSNR and SSIM compared to recently proposed methods such as sRGBFlow and NeCA-W. These experimental results highlight the accuracy of our sRGB noise modeling method and emphasize its applicability in real-world scenarios.

\subsection{Camera-Specific Noise Modeling}
\noindent We demonstrated that our NM-FlowGAN effectively models the noise in the SIDD dataset, which consists of images captured by various cameras, by training a single unified model. However, as mentioned, sRGB noise is significantly affected by hardware specifications such as image sensors. In scenarios where the noise modeling data is captured by cameras with widely varying hardware specifications, employing camera-specific training becomes a suitable solution to improve performance.

In the case of camera-specific training for noise modeling networks, compared to training a unified single model, it may not be necessary to consider the variety of camera types in the noise synthesis process. Instead, training may need to be performed on a small-sized dataset. Therefore, to effectively train noise modeling networks, it is crucial that the noise modeling networks is trained stably even from a small-sized dataset.

Table~\ref{table:table_cam_specific} demonstrates that our proposed NM-FlowGAN shows promising performance even in camera-specific training. This is due to our hybrid architecture, which exploits the advantages of both Normalizing Flows and GAN, allowing our NM-FlowGAN to be stably trained even with small-sized datasets. Specifically, in our pixel-wise noise modeling networks based on Normalizing Flows, stable training can inherently be achieved even on small-sized datasets. Moreover, our GAN-based spatial correlation modeling networks model spatial correlations from synthesized pixel-wise noise, enabling more consistent modeling compared to other GAN-based noise modeling networks. This consistency leads to more stable training.

\newcommand{\cst}{^{*}}
\begin{table}[!t]
\caption{
\textbf{KL divergence of our NM-FlowGAN in camera-specific training. } $\cst$ indicates results obtained from camera-specific training. In camera-specific training, each noise modeling networks is trained using only the data that corresponds to each camera type. 
}
\centering
\renewcommand{\arraystretch}{1.0}
\scalebox{0.85}{
\begin{tabular}{c | ccccc | c }
\toprule
\bf \textsc{Method} & \bf \textsc{G4} & \bf \textsc{GP} & \bf \textsc{IP} & \bf \textsc{N6} & \bf \textsc{S6} & \bf \textsc{ALL} \\
\midrule
\textbf{NM-FlowGAN} & 0.0389 & 0.0259 & 0.0164 & 0.0407 & 0.0334	 & 0.0283 	    \\
\textbf{NM-FlowGAN$\cst$} & 0.0306 & 0.0234 & 0.0103 & 0.0581 & 0.0496 & 0.0318 	\\
\bottomrule
\end{tabular}
}
\label{table:table_cam_specific}
\end{table}

\subsection{Simultaneous Training}
\noindent Our NM-FlowGAN consists of two distinct neural networks: pixel-wise noise modeling networks and spatial correlation networks. These networks are based on Normalizing Flows and GAN, respectively. Our method is capable of training multiple networks simultaneously. However, each network is optimized independently according to its respective loss function, ensuring that the optimization of one network does not affect the other networks. In this paper, we indicate this training strategy to simultaneous training. This implementation is possible because each neural network tends to synthesize a distinct form of noise, and it can improve the training stability and performance of our model in sRGB noise synthesis. Table~\ref{table:table_training_strategy} shows the comparison of noise modeling performance for three different training strategies: the simultaneous training strategy, the joint training strategy without using a stop gradient function, and the two-stage training strategy where the pixel-wise and spatial correlation modeling networks are trained sequentially. 
\begin{table}[!t]
\centering
\caption{
\textbf{Comparison of quantitative performance under different training strategies.} PSNR/SSIM values are measured from the SIDD validation dataset.
}
\setlength{\tabcolsep}{9pt}
\renewcommand{\arraystretch}{1.0}
\scalebox{1.0}{
\begin{tabular}{cccc}
      \toprule
       & \textbf{Joint} & \textbf{Two-Stage} & \textbf{Simultaneous} \\
      \midrule
      KL-Divergence & 0.040 & 0.031 & 0.028 \\
      PSNR/SSIM & 36.47/0.884 & 36.93/0.890 & 37.07/0.891 \\
      \bottomrule
    \end{tabular}}

\label{table:table_training_strategy}
\end{table}

In the case of the two-stage training strategy, it shows comparable performances to our proposed simultaneous training strategy. The two-stage training strategy can also be adopted due to our approach of training the neural networks independently. However, in the case of simultaneous training, our use of simple pixel-wise modeling networks leads to rapid convergence of the negative log likelihood (NLL) loss. As a result, the model is trained in a manner similar to the two-stage training strategy. Therefore, our proposed simultaneous training strategy and the two-stage training strategy show similar performances. In addition, the training time is longer compared to the other training strategy because each neural network needs to be trained sequentially. Therefore, we adopt the simultaneous training strategy for efficiency.

\subsection{Latency}
\begin{table}[!t]
    \caption{\textbf{Average latency of the baselines and our method.}}
  \renewcommand\arraystretch{1.0}
  \centering
  \label{table:table_latency}
  \setlength{\tabcolsep}{10pt}
    \scalebox{0.95}{
    \begin{tabular}{cccc}
      \toprule
       & \textbf{sRGBFlow} & \textbf{NeCA-W} & \textbf{Ours} \\
      \midrule
      Latency (ms) & 31.35 & 106.88 & 83.37 \\
      \bottomrule
    \end{tabular}}
\end{table}

\noindent As shown in Table~\ref{table:table_latency}, our method not only demonstrates better noise synthesis performance compared to NeCA-W, but also has lower latency. Although our method exhibits a higher latency compared to sRGBFlow, the significant performance improvement brings justifies this increase and makes it tolerable in practice. We measured latency using a batch of 16 images, each with a resolution of $256\times256$, on a single NVIDIA A6000 GPU. 

\section{Conclusion}
\label{sec:conclusion}

\noindent In this paper, we introduce a novel hybrid method for modeling sRGB noise. We begin by analyzing the characteristics of sRGB noise and exploring the factors that affect its distribution. From this analysis, we present NM-FlowGAN, noise modeling neural networks that combine two generative models: Normalizing Flows and GAN. By leveraging the strengths of both networks, our method combines pixel-wise noise modeling with Normalizing Flows and spatial correlation modeling with GAN. This combination allows our method to effectively capture noise characteristics and model pixel-to-pixel relationships in sRGB noise. In particular, our approach synthesizes noisy images using only clean images and noise-related factors like camera type or ISO settings, thereby enhancing its applicability in scenarios where paired data is unavailable. Our experimental results show that our method outperforms other baseline methods in sRGB noise modeling and image denoising. Furthermore, we demonstrate the effectiveness of our proposed method and its applicability in real-world scenarios through various experiments.

\bibliographystyle{IEEEtran}
\bibliography{IEEEabrv,nm-flowgan}
 
\vspace{11pt}
\begin{IEEEbiography}[{\includegraphics[width=1in,height=1.25in,clip,keepaspectratio]{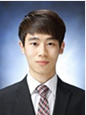}}]{Young~Joo~Han}
received the B.S. degree in computer engineering from Hongik University,
Korea in 2016, and the M.S. degree in computer science from University of Seoul, Korea in 2021.
He is currently pursuing the Ph.D. degree in computer science with University of Seoul. Also, He is currently a researcher with Vieworks. Co., Ltd. His main research interests include medical image processing and low-level computer vision.
\end{IEEEbiography}

\vspace{11pt}
\begin{IEEEbiography}[{\includegraphics[width=1in,height=1.25in,clip,keepaspectratio]{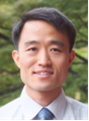}}]{Ha-Jin~Yu}
received the B.S., M.S. and Ph.D. degrees in computer science from KAIST, Korea, in 1990, 1992, 1997, respectively. From 1997 to 2000, he was a Senior Researcher with LG Electronics. From 2000 to 2002, he was a Director with SL2 Ltd. Since 2002, he has been a Professor with the Computer Science Department, University of Seoul. His research interests include the speech and speaker recognition, acoustic scene classification and machine learning. He is an Editor of The Journal of the Acoustical Society of Korea and the Journal of the Korean Society of Speech Sciences.
\end{IEEEbiography}

\vfill

\end{document}